\documentclass[runningheads]{llncs}
\usepackage{xcolor}
\usepackage[T1]{fontenc}
\usepackage{graphicx}
\usepackage{amsmath}
\usepackage{amssymb}
\usepackage{makecell}
\usepackage{booktabs}
\usepackage{tabularx}
\usepackage{array}

\newcolumntype{Y}{>{\centering\arraybackslash}X}
\begin{document}
\title{Selective Spatio-Temporal Fruit Inspector (S$^2$FI): A Mamba-based Framework for Rotational Fruit Quality Assessment}
\author{
Hao Tian\inst{1}\orcidID{0009-0007-1446-4083} \and
Jie Hu\inst{1}\orcidID{0009-0003-7473-4226} \and
Guoqi Shan\inst{1}\orcidID{0009--0005-7825-4064}\and Fan Wang\inst{1}\orcidID{0009--0001-5507-6956}\and Wenzhuo Zhang\inst{2}\orcidID{0009--0000-5849-489X}\and Youshan Zhang\inst{1**}\orcidID{0000--0002-0074-0979}
}
\authorrunning{Tian.H et al.}
\institute{School of Artificial Intelligence, Chuzhou University, Anhui, China \and
School of Computer and Information Engineering, Chuzhou University, China\\
$^{**}$Corresponding Author, youshan\_zhang@chzu.edu.cn \thanks{This research was funded by the Research Projects of Anhui Province Education Department (Grant No. 2025AHGXZK20009).}}
\maketitle
\begin{abstract}
The global fruit and vegetable industry confronts severe post-harvest sorting challenges for small and medium-scale growers. Large intelligent sorting systems are too costly, while manual sorting suffers from low efficiency, inconsistent standards, and high labor demand. To bridge this gap between expensive equipment and inefficient manual work, this study develops a low-cost, high-efficiency, lightweight fruit sorting system driven by deep learning, suitable for small and medium growers. We propose the Selective Spatio-Temporal Fruit Inspector (S$^2$FI) architecture, which combines a 3D convolutional encoder with a Mamba-based backbone.It efficiently extracts spatiotemporal features, adaptively weighs informative frames during fruit rotation, and models long-range dependencies with linear complexity, reducing motion blur and occlusion. The hardware uses a dual-conveyor and curved rail design, supporting household power, fast deployment, and lightweight embedded deployment to improve affordability. We also built a conveyor-based fruit quality video dataset with 6,460 samples. Experiments on the held-out test set show that our method achieves the best trade-off
among accuracy, efficiency, and practicality, outperforming the compared baselines on the test set, achieving an excellent balance of accuracy, speed, and practicality for real-world sorting.

\end{abstract}

\section{Introduction}

Driven by population expansion and consumption upgrade, the global fruit industry has maintained steady development. Nevertheless, post-harvest sorting has long been a critical constraint restricting industrial progress. Small and medium-sized growers, cooperatives, and other small-scale producers have long faced a dilemma in the post-harvest sorting process: large intelligent sorting systems, which often cost millions of dollars, are too expensive to deploy, whereas conventional manual sorting suffers from low efficiency, high labor costs, and unstable sorting standards. Such a mismatch between market demand and available solutions has highlighted the strong market demand for low-cost, lightweight intelligent sorting systems adapted to small-scale operation scenarios.

Prior research on automated fruit sorting has established essential groundwork, though significant limitations in practical scenario adaptation persist. Ani et al. \cite{ani2018image} developed a MATLAB-based tomato grading approach, yet it suffered from a 20-second per-image processing delay—an insurmountable barrier to large-scale application. Oyefeso et al. \cite{oyefeso2025automated} proposed an image processing combined with Support Vector Machine (SVM) system, which achieved 80–100\% accuracy across multiple fruit types but failed to adapt to complex defect detection scenarios. Chen et al. \cite{chen2021deep} designed a Convolutional Neural Network (CNN)-based citrus sorting system with 93.6\% defect detection precision, yet it required specialized technical deployment. Two core challenges in post-harvest sorting remain unresolved: first, large sorting equipment places an excessive resource burden on small operators; second, lightweight equipment lacks sufficient multi-dimensional detection accuracy, cannot effectively simulate the spatiotemporal characteristics of rotating fruit sequences, and cannot selectively focus on keyframes. In addition, most existing solutions require specialized operation and industrial power supplies, making them incompatible with grassroots agricultural application scenarios.

To tackle the above bottlenecks, our contributions are three-fold:
\begin{itemize}
    \item We construct a dedicated fruit video classification dataset named \textit{Conveyor-based Fruit Quality Assessment Dataset}, which contains 6460 video clips that conform to real agricultural scenarios and provides a benchmark for evaluating fruit quality classification models.
    \item We propose the \textit{Selective Spatio-Temporal Fruit Inspector (S\textsuperscript{2}FI)} framework, which integrates a 3D convolution encoder and Mamba-based selective state-space modeling to extract spatiotemporal features with linear complexity $O(L)$, enabling accurate multi-species sorting and lightweight embedded deployment.
    \item We design a compact, low-cost hardware system with dual-conveyor and curved-rail mechanisms that supports household power and on-site deployment, filling the gap of affordable high-precision sorting equipment for small-scale producers.
\end{itemize}

\section{Related Work}
\textbf{Traditional Image Processing and Sensor-based Methods}. Classical fruit sorting methods extracted handcrafted visual features via thresholding, contour analysis, and rule-based classification under controlled lighting, and non-destructive sensing techniques like visible–near infrared spectroscopy and bioimp-edance were also applied for maturity detection—yet O’Brien and Alamar \cite{o2025overview} identified that both approaches demanded specialized hardware and careful calibration. Though simple and fast, these methods lacked robustness in real agricultural settings due to sensitivity to illumination, moisture, and morphological variation, and Cubero et al. \cite{cubero2011advances} further noted that the absence of semantic understanding hindered discrimination of visually similar conditions, limiting their applicability beyond labs.
Existing CNN‑Mamba video frameworks typically adopt a two‑stage paradigm: extract spatial feature maps via CNN backbone, then feed flattened spatial tokens into sequential Mamba blocks for temporal modeling. Such schemes treat spatial‑statistical cues and temporal dynamics independently, and ignore fine‑grained intra‑frame statistical information which is critical for fruit defect grading. Different from them, our S$^2$FI introduces multi‑statistical spatial embedding branch. Besides conventional CNN spatial feature and Mamba temporal modeling, it explicitly extracts first‑order and second‑order statistical features from video frames. The statistical branch cooperates with CNN appearance stream and Mamba temporal stream, so that local visual appearance, global statistical distribution and frame‑level temporal evolution can be jointly modeled for low‑resolution fine‑grained video classification.

\textbf{Machine Learning (ML)-based Grading Methods}. Shallow ML models addressed rule-based rigidity using engineered features. SVMs \cite{cortes1995support} gained popularity for effectiveness with limited data. Wang et al. \cite{wang2023fruit} integrated mechanical pressure and visual cues for joint appearance-firmness assessment. Azarmdel et al. \cite{azarmdel2020evaluation} combined image processing with ANN and SVM for mulberry ripeness grading, achieving high accuracy under controlled conditions. Multimodal fusion strategies—e.g., integrating VIS–NIR spectroscopy with bioimpedance—further improved accuracy, as demonstrated by Liu et al. \cite{liu2025multimodal} for kiwifruit maturity grading using SVM with fused spectral-impedance features. Yet these approaches still relied on handcrafted feature concatenation and lacked end-to-end representation learning. Performance remained constrained by manual feature design—especially for heterogeneous defects or cross-species tasks—and lacked efficient long-sequence modeling or dynamic focus on critical rotation frames, impairing real-time sorting.

\textbf{Deep Learning-based Sorting Methods}. Deep learning enables end-to-end visual feature learning, with CNNs dominating fruit quality inspection \cite{rizzo2023fruit}. However, standard CNNs lack temporal modeling for rotation-based sorting. 3D CNN variants (e.g., C3D, I3D, X3D \cite{feichtenhofer2020x3dexpandingarchitecturesefficient}) capture spatiotemporal features but struggle with long-range dependencies due to limited receptive fields. Attention-enhanced architectures address this: Tuteja et al. \cite{tuteja2025cnnattlstm} proposed CNNAttLSTM, achieving 99\% accuracy with only 3.7M parameters for jackfruit disease classification. ViViT \cite{arnab2021vivit} and VideoMAE \cite{tong2022videomae} extend Transformers to video but suffer from high complexity or lack of selective frame focus. Hybrid CNN–Transformer models like ResNet-ViT-SVM \cite{sabanci2025resnet} balance accuracy and efficiency (99.40\% on date fruit), yet remain static-image based. Mamba provides linear complexity (O(L)) for long-sequence modeling; Res-VMamba \cite{chen2024res} shows promise in food classification, but its integration with spatial feature extraction and multi-statistical fusion is unexplored in fruit sorting. Moreover, most frameworks rely solely on RGB inputs, focus narrowly on visible defects, and prioritize accuracy over efficiency—unsuitable for low-cost edge devices on small farms.

\textbf{Lightweight Intelligent Sorting Systems}. Recent efforts toward edge-deployable fruit sorting follow two main directions: efficient model design and multi-view imaging strategies. For efficient model design, techniques such as network pruning \cite{han2015learning} and quantization \cite{courbariaux2015binaryconnect} are widely adopted. Cheng et al. \cite{cheng2021lightweight} proposed a compact CNN detector for real-time fruit localization, while Albahli \cite{albahli2025agrifusionnet} systematically reviewed lightweight deep learning models and multisource fusion strategies for agricultural plant disease diagnosis, highlighting the potential of edge-deployable architectures in real-time crop inspection tasks. Chen et al. \cite{chen2026real} developed YOLO-BFD, a YOLOv7-tiny-based lightweight detector for banana freshness grading, achieving 97.8\% accuracy at 61 FPS on Jetson Orin NX with 41.9\% parameter reduction, and explicitly validated mechanical arm rotation for multi-angle inspection. For overlapping or touching fruits, Rani et al. \cite{rani2024topogeofusion} proposed TopoGeoFusion, a geometry–topology feature fusion method achieving 100\% accuracy on gooseberry classification from smartphone RGB images without depth sensors. Despite these advances, current solutions prioritize single-aspect optimization over comprehensive evaluation—efficient detectors focus on speed, while multi-view or geometry-based methods target specific imaging challenges. Neither line of work integrates multi-dimensional quality assessment, robust spatiotemporal modeling of rotation sequences, selective attention, or multi-statistical fusion, leaving a critical gap that this work addresses.

In summary, existing methods fail to balance lightweight deployment, multi-dimensional grading, robust spatiotemporal modeling, selective attention, and accuracy. 
\section{Methodology}
\label{sec:methodology}

This section comprehensively details the key components of the proposed \textbf{S\textsuperscript{2}FI} model, including selective state space modeling, adaptive loss function, and multi-statistical fusion mechanisms, all of which are specifically designed to conduct thorough fruit quality evaluation based on rotation sequences.

\begin{figure}[htbp]
    \centering
    \includegraphics[width=\linewidth]{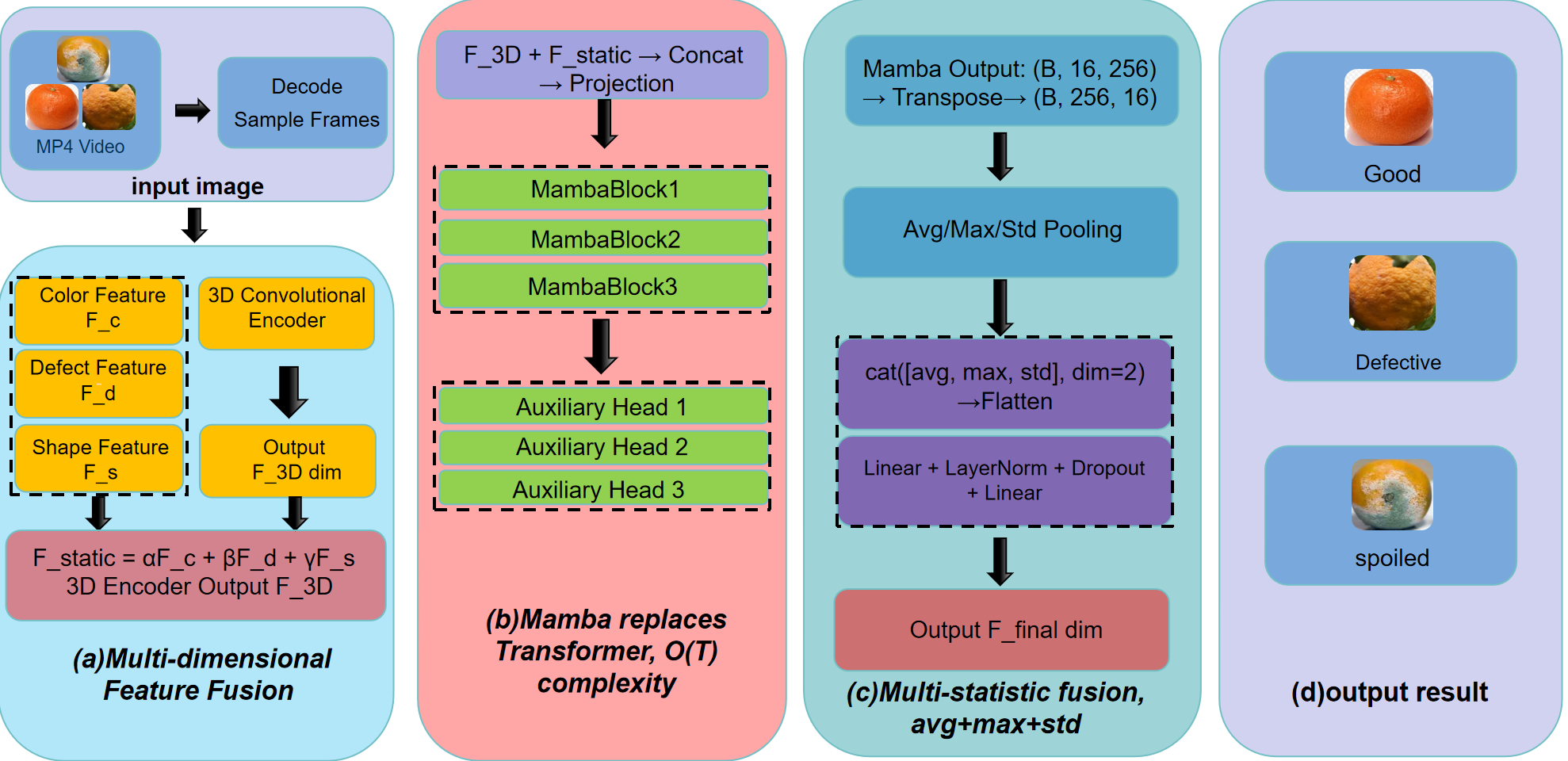}
    \caption{Architecture of the proposed S$^2$FI model. From left to right: (a) multi-dimensional feature fusion via color, defect, and shape cues from 3D convolutional encoder; (b) Mamba blocks with auxiliary supervision heads for deep temporal modeling; (c) multi-statistic fusion (avg/max/std) to aggregate discriminative temporal patterns; (d) final classification outputting fruit quality categories.}
    \label{fig:architecture}
\end{figure}

\subsection{Input Representation and Spatiotemporal Feature Encoding}
\subsubsection{Input Format}

The model takes as input video segments of rotating fruits, which are formatted as follows:
\begin{equation}
    \mathbf{X} \in \mathbb{R}^{B \times C \times T \times H \times W}
\end{equation}
In this formulation, $B$ denotes the batch size, $C=3$ corresponds to RGB color channels, $T=16$ represents the number of temporal frames, and $H=W=112$ indicates the spatial resolution of each frame. All input pixel values are normalized to the range of $[0,1]$.

\subsubsection{3D Convolutional Encoder}

A 3D convolutional encoder denoted as $\mathcal{E}$ is used to extract hierarchical spatiotemporal features from the input video clips, following the formulation:
\begin{equation}
    \mathbf{F}_t = \mathcal{E}(\mathbf{X}) = \text{Conv3D}_{\theta_e}(\mathbf{X})
\end{equation}
Here, $\mathbf{F}_t \in \mathbb{R}^{B \times C' \times T' \times H' \times W'}$ is the resulting feature tensor, where $C'=128$ is the number of feature channels, $T'$ is the number of temporal frames retained, and $H'=W'=4$ are the spatial dimensions after adaptive pooling. The encoder implements a three-stage progressive feature extraction process:
\begin{equation}
\begin{split}
\mathbf{F}^{(1)} &= \sigma(\text{BN}(\text{Conv}_{3\times5\times5}(\mathbf{X}))) \\
\mathbf{F}^{(2)} &= \sigma(\text{BN}(\text{Conv}_{3\times3\times3}(\mathbf{F}^{(1)}))) \\
\mathbf{F}^{(3)} &= \text{AdaptiveAvgPool}(\sigma(\text{BN}(\text{Conv}_{3\times3\times3}(\mathbf{F}^{(2)}))))
\end{split}
\end{equation}
where $\sigma(\cdot)$ stands for the ReLU activation function, and BN refers to the batch normalization operation.
These three branches are not simple stacked components, and there exists explicit synergy effect among them.
The CNN appearance branch captures local texture and defect visual cues of fruit frames, yet it is vulnerable to illumination change and fruit rotation noise.
The multi‑statistical embedding branch supplements global first‑/second‑order distribution information, compensating the instability of local CNN features under complex real‑world disturbance.
The Mamba temporal branch further models frame‑wise temporal evolution across video clips, eliminating frame‑level ambiguity by leveraging information from adjacent frames.
Feature fusion aggregates three types of complementary representations. CNN provides fine local visual evidence, statistical branch enhances robustness against noisy samples, and Mamba captures temporal context; their joint representation mitigates training collapse caused by dataset imbalance, and finally yields discriminative embedding for fruit quality grading. Removing any one branch will break this complementary mechanism and degrade overall performance.

\subsection{Selective State Space Modeling}
\subsubsection{Feature Projection}

The extracted spatiotemporal features are first flattened and then projected into a unified embedding space using the following transformation:
\begin{equation}
    \mathbf{Z}_t = \mathcal{P}(\text{Reshape}(\mathbf{F}_t)) = \text{Dropout}(\text{LayerNorm}(\mathbf{W}_p\mathbf{f}_t + \mathbf{b}_p))
\end{equation}
In this equation, $\mathbf{Z}_t \in \mathbb{R}^{B \times T' \times d}$ denotes the projected feature sequence with the embedding dimension $d=256$, and $\mathbf{W}_p \in \mathbb{R}^{d \times (C' \cdot H' \cdot W')}$ is the learnable projection matrix.

\subsubsection{Selective State Space\cite{gu2024mamba}}

A key innovation of the S$^2$FI model is its selective state space modeling mechanism. In contrast to traditional attention-based methods with $O(L^2)$ computational complexity, our approach achieves linear complexity $O(L)$ by leveraging parameterized state transition operations:
\begin{equation}
\begin{split}
\mathbf{x}_t &= \mathbf{W}_{\text{in}}[\mathbf{z}_t\\
\tilde{\mathbf{x}}_t &= \text{SiLU}(\text{Conv1D}(\mathbf{x}_t)) \\
\mathbf{h}_t &= \mathbf{A}\mathbf{h}_{t-1} + \mathbf{B}\tilde{\mathbf{x}}_t \\
\mathbf{y}_t &= \mathbf{C}\mathbf{h}_t + \mathbf{D}\tilde{\mathbf{x}}_t
\end{split}
\end{equation}
Here, $\mathbf{A} = -\exp(\mathbf{A}_{\log}) \in \mathbb{R}^{d \times N}$ is the state transition matrix (where $N=16$ indicates the dimension of the state space), $\mathbf{B}, \mathbf{C} \in \mathbb{R}^{d \times N}$ are the input and output projection matrices respectively, and $\mathbf{D}$ is a parameter for the skip connection. The selective mechanism is defined as:
\begin{equation}
    \mathbf{A}_{\log} = \log(\text{arange}(1, N+1)) \otimes \mathbf{1}_d
\end{equation}
In practice, the matrices $\mathbf{B}$, $\mathbf{C}$ and the step parameter $\Delta$
are generated from the input $\mathbf{z}_t$ through small linear projections, which
makes the state-space dynamics input-dependent. This input-dependent parameterization
allows the model to adaptively emphasize frames whose content is informative for
quality assessment, while the state-space formulation keeps the computational cost
linear in the sequence length.

\subsubsection{Temporal Sequence Processing}

The full temporal modeling pipeline utilizes a stack of selective state space blocks, formulated as:
Each block maintains the original temporal resolution and is equipped with a residual connection, i.e.,
\begin{equation}
\mathbf{H}^{(l)} = \mathbf{H}^{(l-1)} + \mathcal{M}^{(l)}(\mathbf{H}^{(l-1)}),
\end{equation}
where $\mathcal{M}^{(l)}$ represents the transformation computed by the $l$-th Mamba block.

In this equation, $\mathcal{M}^{(k)}$ represents the $k$-th Mamba transformation module.

\subsection{Multi-Statistical Feature Fusion}

The S$^2$FI model incorporates a comprehensive multi-statistical fusion strategy specifically tailored for video sequences of rotating fruits. Based on the final temporal sequence representation $\mathbf{H} = \mathbf{H}^{(4)} \in \mathbb{R}^{B \times T' \times d}$, three complementary statistical indicators are calculated to capture diverse feature characteristics:
\begin{equation}
\boldsymbol{\mu} = \frac{1}{T'}\sum_{t=1}^{T'} \mathbf{h}_t \quad,
\boldsymbol{\phi} = \max_{t} \mathbf{h}_t \quad,
\boldsymbol{\sigma} = \sqrt{\frac{1}{T'}\sum_{t=1}^{T'} (\mathbf{h}_t - \boldsymbol{\mu})^2} \quad 
\end{equation}
These statistical metrics reflect distinct aspects of fruit quality during the rotation process: temporal mean $\boldsymbol{\mu}$ captures the overall feature characteristics of the fruit, temporal maximum $\boldsymbol{\phi}$ emphasizes prominent defect areas that affect quality assessment, and temporal standard deviation $\boldsymbol{\sigma}$ quantifies the consistency of fruit appearance throughout the rotation sequence. The final fused feature representation is obtained through the following formulation:
\begin{equation}
    \mathbf{f}_{\text{fused}} = \mathbf{W}_f[\boldsymbol{\mu}, \boldsymbol{\phi}, \boldsymbol{\sigma}] + \mathbf{b}_f
\end{equation}
Here, $\mathbf{W}_f \in \mathbb{R}^{d \times 3d}$ denotes the learnable fusion weight matrix, and the notation $[\cdot,\cdot,\cdot]$ stands for the feature concatenation operation.

\subsection{Super Adaptive Composite Loss Function}

The S$^2$FI model adopts a novel Super Adaptive Composite Loss (SACL) function, which dynamically adjusts its parameters according to sample difficulty and class distribution characteristics:
\begin{equation}
    \mathcal{L}_{\text{SACL}} = \frac{1}{B}\sum_{i=1}^{B} w_i \cdot g_i \cdot \left[ \alpha \cdot \mathcal{L}_{\text{focal}}(x_i,y_i) + \beta \cdot \mathcal{L}_{\text{smooth}}(x_i,y_i) \right]
    \label{eq:sacl}
\end{equation}
In this formula, $x_i$ and $y_i$ represent the input feature and corresponding target label of the $i$-th sample, respectively; $\alpha$ and $\beta$ are learnable weight parameters that balance the two loss components, and $g_i$ is a gradient weight that emphasizes hard samples (defined below). The specific formulations of the focal loss and smooth loss are given as follows:
\begin{equation}
    \mathcal{L}_{\text{focal}}(x_i,y_i) = -(1-p_{i,y_i})^{\gamma} \log(p_{i,y_i})
\end{equation}
\begin{equation}
    \mathcal{L}_{\text{smooth}}(x_i,y_i) = -\sum_{c=1}^{K} \tilde{y}_{i,c} \log(p_{i,c})
\end{equation}
The class weight $w_i$ integrates both the inverse class frequency and the magnitude of the running loss, which is defined as:
\begin{equation}
    w_i = \sqrt{\frac{1}{N_{y_i}}} \cdot \frac{1}{\ell_{y_i} + \epsilon}
\end{equation}
Here, $N_c$ denotes the number of samples in class $c$, and $\ell_c$ stands for the accumulated running loss of class $c$ during training. The adaptive focusing parameter $\gamma$ is adjusted dynamically based on the difficulty level of each sample:
\begin{equation}
    \gamma = \text{clamp}\left(\gamma_0 + \lambda \cdot d_i,\ \gamma_{\min},\ \gamma_{\max}\right)
\end{equation}
where the sample difficulty is defined as $d_i = 1 - p_{i,y_i}$, i.e., the complement of the predicted probability of the true class (larger $d_i$ indicates a harder sample), and $[\gamma_{\min}, \gamma_{\max}]$ bounds the focusing parameter. To emphasize hard-to-classify samples, a gradient weight is introduced, formulated as:
\begin{equation}
    g_i = \exp\left(\frac{\ell_i - \mu_{\ell}}{\sigma_{\ell}} \cdot \tau\right)
\end{equation}
which is multiplied into the per-sample loss in Eq.~(\ref{eq:sacl}); here, $\mu_{\ell}$ and $\sigma_{\ell}$ are the running mean and running standard deviation of the loss values across training batches, respectively. The label smoothing distribution, which mitigates overfitting, is defined as:
\begin{equation}
    \tilde{y}_{i,c} = 
    \begin{cases} 
        1-\epsilon & \text{if } c=y_i \\
        \dfrac{\epsilon}{K-1} & \text{otherwise}
    \end{cases}
\end{equation}

In the above equations, $\mu_{\ell}$ and $\sigma_{\ell}$ are the running mean and running standard deviation of the loss values across training batches, respectively, and $\epsilon$ is a hyperparameter that is adaptively adjusted throughout the training process.

\section{Experiment}
\subsection{Dataset Description}

To effectively evaluate the performance of the proposed fruit sorting system in practical scenarios, we established a specialized dataset that accurately mirrors real-world agricultural working environments. This dataset was acquired through on-site data collection at modern integrated fruit industrial parks, where a custom-developed experimental platform was deployed to capture video sequences of fruits under standardized control conditions. The dataset is divided into training, validation, and test subsets in a ratio of 7:1:2, with each subset covering three distinct categories of fruit quality. To prevent information leakage, the dataset is split at the video level: all clips
recorded from the same physical fruit, the same rotation session, and the same
acquisition batch are assigned to the same subset, so that adjacent frames or
segments from the same fruit never appear in different splits.The specific distribution of samples among these quality categories and data subsets is shown in Table \ref{tab:dataset_summary}.

\begin{table}[htpb]
\centering
\caption{Summary of the Fruit Sorting Dataset}
\label{tab:dataset_summary}
\begin{tabular}{|l|c|c|c|c|}
\hline
\textbf{Category} & \textbf{Training Set} & \textbf{Validation Set} & \textbf{Test Set} & \textbf{Total} \\
\hline
Good Fruit        & 2729                  & 390                     & 779               & 3898           \\
Substandard Fruit & 1441                  & 206                     & 411               & 2058           \\
Rotten Fruit      & 353                   & 50                      & 101               & 504            \\
\hline
\textbf{Total}    & \textbf{4523}         & \textbf{646}            & \textbf{1291}     & \textbf{6460}   \\
\hline
\end{tabular}
\end{table}

\begin{figure}[htbp]
    \centering
    \includegraphics[width=\textwidth]{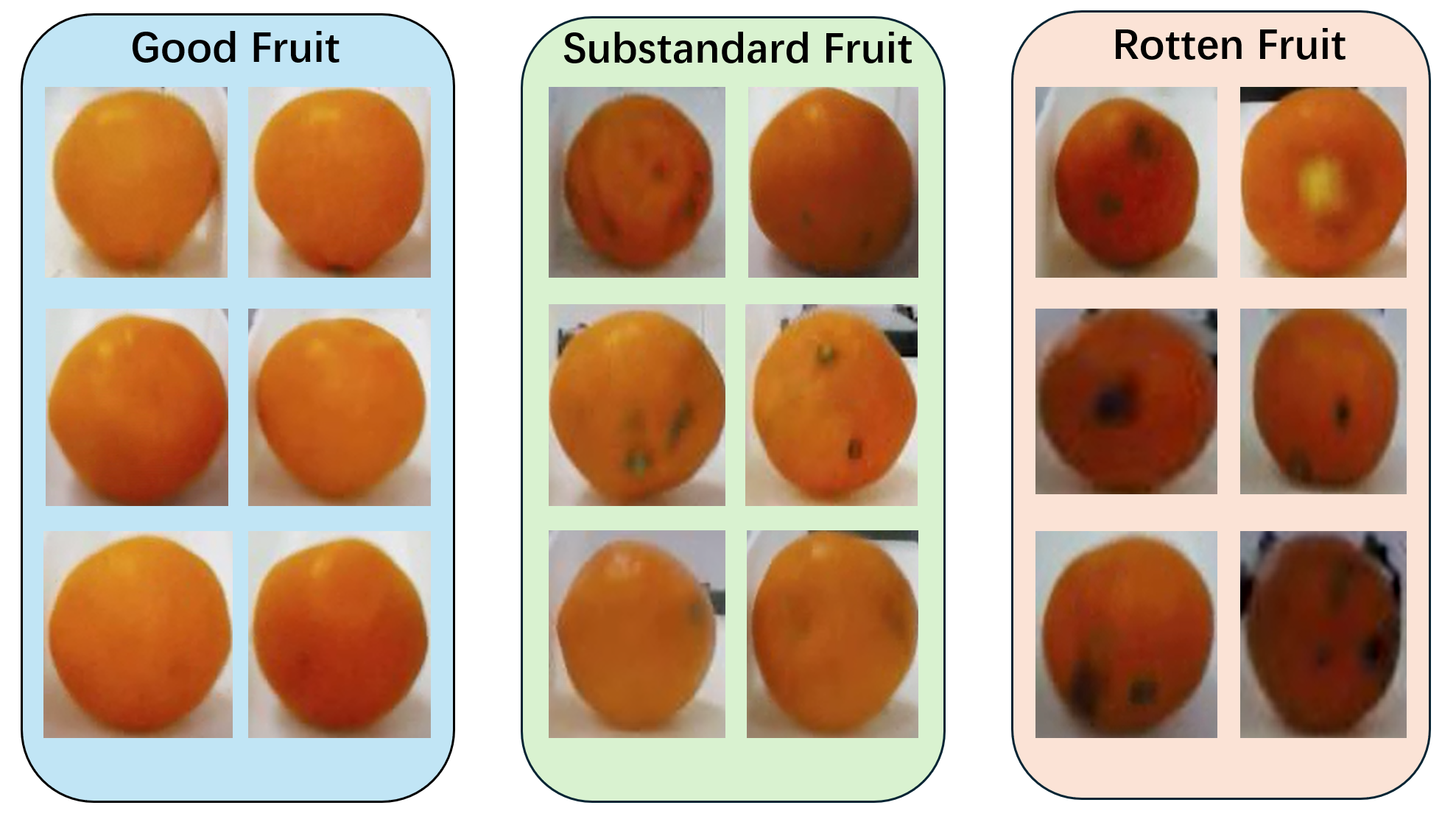}
    \caption{Fruit Morphological Classification Standards}
    \label{fig2}
\end{figure}

The detailed morphological criteria for sound fruit, substandard fruit, and rotten fruit are illustrated in Figure \ref{fig2}. Each fruit image in this figure is taken from one of the 6 frames in the video.

\subsection{Implementation Details}

All experiments were conducted on a computing node with an RTX 5060 Ti 16G using PyTorch and CUDA acceleration.
Eight baseline models (C3D, I3D, R3D, MC3D, Transformer, DVFL-Net, MoCLIP-Lite, X3D) were evaluated. Key hyperparameters: initial learning rate 0.00005 (AdamW optimizer), batch size 8, 100 training epochs.The proposed Super Adaptive Composite Loss (SACL) was used to handle class imbalance.
 Input video clips were resized to 16 frames of 112$\times$112 pixels and normalized to [0, 1]. Following standard practice for imbalanced classification, models were assessed via
the macro-averaged F1 score together with the Mean Class Accuracy (MCA), i.e.,
the arithmetic mean of the per-class accuracies, which is reported for the three
fruit grades (high-quality, substandard, defective) as well.
All hyper-parameters were tuned on the training and validation splits only;
the test set was evaluated once with the frozen configuration, and the five
test-set runs were used solely to estimate the uncertainty of the reported metrics.

\textbf{Baseline implementation.} For all compared baseline models, we maintain identical input resolution (16 frames, $112\times112$), dataset split, and optimization pipeline for fair comparison. We adopt official public pre‑trained weights where available and replace the final classification head for our three‑class fruit quality task. We perform hyper-parameter search over learning rate, batch size, and weight decay, and select the best configuration based on validation performance.
 Even with sufficient tuning, several baselines (C3D, vanilla Transformer, MoCLIP‑Lite) still suffer from training collapse on this imbalanced fine‑grained fruit dataset. These architectures tend to over‑fit to the dominant majority class under severe label imbalance, failing to capture subtle visual differences among fine‑grained defective fruit samples.
\subsection{Results}

\begin{table}[tbp]
\centering
\caption{Validation performance comparison of fruit quality classification models.Note: Medium corresponds to substandard fruit; Bad corresponds to rotten fruit.Some baseline models suffer from training collapse, tending to predict the majority class and yielding near-zero accuracy on minority categories.}

\label{tab:model_perf}

\setlength{\tabcolsep}{3pt}
\renewcommand{\arraystretch}{1.05}

\resizebox{\linewidth}{!}{%
\begin{tabular}{|l|c|c|c|c|c|c|}
\hline
\textbf{Model} &
\makecell{\textbf{Good}\\\textbf{(\%)}} &
\makecell{\textbf{Medium}\\\textbf{(\%)}} &
\makecell{\textbf{Bad}\\\textbf{(\%)}} &
\makecell{\textbf{MCA}\\\textbf{(\%)}} &
\textbf{Macro F1} &
\makecell{\textbf{Params}\\\textbf{(M)}} \\
\hline
R3D \cite{tran2018closer} & 24.1 & 75.2 & 100.0 & 66.5 & 0.608 & 900.7 \\
I3D \cite{carreira2017quo} & 45.4 & 62.1 & 100.0 & 69.2 & 0.672 & 529.2 \\
C3D \cite{tran2015learning} & 100.0 & 0.0 & 0.0 & 33.3 & 0.251 & 73.8 \\
Transformer \cite{vaswani2017attention} & 100.0 & 0.0 & 0.0 & 33.3 & 0.251 & 274.9 \\
MoCLIP-Lite \cite{huang2025moclipliteefficientvideorecognition}
& 0.0 & 0.0 & 100.0 & 33.3 & 0.2084 & 28.5 \\
DVFL-Net \cite{ullah2025dvfl} & 67.2 & 87.4 & 100.0 & 84.9 & 0.824 & 32.7 \\
X3D \cite{feichtenhofer2020x3d} & 7.4 & 58.3 & 66.0 & 43.9 & 0.572 & 6.7 \\
MC3D \cite{zhou2018mict} & 0.3 & 99.0 & 100.0 & 66.4 & 0.499 & 2451.8 \\
\textbf{S$^2$FI (Ours)} & \textbf{71.5} & \textbf{28.2} &
\textbf{100.0} & \textbf{66.6} & \textbf{0.664} & \textbf{36.6} \\
\hline
\end{tabular}%
}

\vspace{2pt}
\begin{minipage}{\linewidth}
\footnotesize
\raggedright
Params: trainable parameters in millions.\\
\end{minipage}
\end{table}

\begin{table}[tbp]
\centering
\caption{Testing performance comparison of fruit quality classification models.
Note: Medium corresponds to substandard fruit; Bad corresponds to rotten fruit.
Some baseline models suffer from training collapse, tending to predict the majority
class and yielding near-zero accuracy on minority categories.}

\label{tab:model_test_perf}

\setlength{\tabcolsep}{3pt}
\renewcommand{\arraystretch}{1.05}

\resizebox{\linewidth}{!}{%
\begin{tabular}{|l|c|c|c|c|c|c|}
\hline
\textbf{Model} &
\makecell{\textbf{Good}\\\textbf{(\%)}} &
\makecell{\textbf{Medium}\\\textbf{(\%)}} &
\makecell{\textbf{Bad}\\\textbf{(\%)}} &
\makecell{\textbf{MCA}\\\textbf{(\%)}} &
\textbf{Macro F1} &
\makecell{\textbf{Params}\\\textbf{(M)}} \\
\hline
R3D \cite{tran2018closer} & 66.1 & 34.9 & 1.0 & 67.0 & 0.589 & 900.7 \\
I3D \cite{carreira2017quo} & 65.5 & 35.2 & 96.8 & 65.8 & 0.632 & 529.2 \\
C3D \cite{tran2015learning} & 100.0 & 0.0 & 0.0 & 33.3 & 0.251 & 73.8 \\
Transformer \cite{vaswani2017attention} & 100.0 & 0.0 & 0.0 & 33.3 & 0.251 & 274.9 \\
MoCLIP-Lite \cite{huang2025moclipliteefficientvideorecognition}
& 0.07 & 36.9 & 14.0 & 26.3 & 0.235 & 28.5 \\
DVFL-Net \cite{ullah2025dvfl} & 55.5 & 20.7 & 43.7 & 40.6 & 0.424 & 32.7 \\
X3D \cite{feichtenhofer2020x3d} & 50.2 & 35.4 & 31.0 & 38.3 & 0.396 & 6.7 \\
MC3D \cite{zhou2018mict} & 12.5 & 31.6 & 24.2 & 22.8 & 0.273 & 2451.8 \\
\textbf{S$^2$FI (Ours)} & \textbf{71.3} & \textbf{48.1} &
\textbf{100.0} & \textbf{73.2} & \textbf{0.723} & \textbf{36.6} \\
\hline
\end{tabular}%
}

\vspace{2pt}
\begin{minipage}{\linewidth}
\footnotesize
\raggedright
Params: trainable parameters in millions.\\
\end{minipage}
\end{table}
On the validation set, S$^2$FI obtains an MCA of 66.6\% and a Macro F1 score of 0.664, while DVFL-Net achieves a higher validation MCA and Macro F1.
Since the validation set is used for model selection, we treat the held-out test set as the primary basis for comparing generalization performance.

In Table \ref{tab:model_test_perf}, S$^2$FI exhibits strong generalization capability, achieving the highest MCA of 73.2\% and a Macro F1 score of 0.723
on the unseen test data.

The accuracy for the medium-grade category on the validation set is 28.2\%,
which may reflect the higher ambiguity of the samples in that split; both splits are reported for completeness, and the held-out test set is used as the primary benchmark.

It maintains 100\% accuracy for ``Bad'' fruits and significantly outperforms other methods on the challenging ``Medium'' category, with an accuracy of 48.1\% on the test set. In contrast, traditional 3D CNNs are hindered by class imbalance issues, while Transformer and MC3D models fail to translate their high computational complexity into meaningful performance improvements. These results confirm that S$^2$FI strikes an optimal balance among classification accuracy, computational efficiency, and generalization ability, making it well‑suited for real‑world fruit sorting applications.

\subsection{Ablation Study}
To systematically assess the contribution of each component in the proposed
$S^2$FI framework, we conduct ablation experiments under a common training
protocol, data partition, and evaluation procedure. Five components are
examined: the 3D convolutional encoder, the Mamba-based backbone, the
multi-statistical feature fusion branch, the auxiliary supervision head, and
the SACL loss. The corresponding results are reported in
Table~\ref{tab:ablation}.

Removing the 3D encoder produces the most substantial degradation, with the
Medium recall falling to $0.3\%$ and the Macro-F1 score to $0.389$. This
result highlights the importance of spatio-temporal representation learning
for video-based fruit inspection. Replacing the Mamba backbone reduces the
Medium recall to $3.9\%$ and the Macro-F1 score to $0.587$, demonstrating the
benefit of long-range temporal modeling. Eliminating the multi-statistical
fusion branch further lowers the MCA and weakens recognition of Medium
samples, indicating that first- and second-order statistics provide
complementary information under rotation and illumination variations.
Removing the auxiliary supervision head yields a relatively higher MCA among
the ablated variants, mainly owing to its improved Medium recall; however,
its Macro-F1 remains below that of the complete model. Finally, replacing
SACL with focal loss leads to lower MCA and Macro-F1 scores, supporting the
use of adaptive loss weighting for imbalanced fruit-quality classification.
Overall, the complete $S^2$FI model achieves the best overall performance,
with an MCA of $73.2\%$ and a Macro-F1 score of $0.723$, confirming the
effectiveness of jointly incorporating all proposed components.
\begin{table}[tbp]
\centering
\scriptsize
\setlength{\tabcolsep}{2pt}
\renewcommand{\arraystretch}{0.95}

\caption{Ablation study of individual components in $S^2$FI.}
\label{tab:ablation}

\begin{tabularx}{0.90\linewidth}{@{}
>{\raggedright\arraybackslash}X
YYYYY
@{}}
\toprule
\textbf{Variant} &
\makecell[c]{\textbf{Good}\\\textbf{(\%)}} &
\makecell[c]{\textbf{Medium}\\\textbf{(\%)}} &
\makecell[c]{\textbf{Bad}\\\textbf{(\%)}} &
\makecell[c]{\textbf{MCA}\\\textbf{(\%)}} &
\textbf{Macro F1} \\
\midrule

w/o 3D Encoder
& 74.2 & 0.3 & 99.8 & 58.1 & 0.389 \\

w/o Mamba Backbone
& 98.3 & 3.9 & 99.8 & 67.3 & 0.587 \\

w/o Multi-stat. Fusion
& 96.4 & 12.9 & 100.0 & 69.7 & 0.623 \\

w/o Auxiliary Head
& 87.9 & 24.5 & 100.0 & 70.8 & 0.601 \\

w/o SACL
& 97.7 & 2.5 & 93.9 & 64.7 & 0.563 \\

\midrule

\textbf{Full $S^2$FI}
& 71.3 & 48.1 & 100.0 & 73.2 & 0.723 \\

\bottomrule
\end{tabularx}
\end{table}

\section{Discussion}
Our model achieves 100\% defective fruit detection, outperforming I3D by 3.2\% to prevent spoilage; Its substandard fruit accuracy is 48.1\% on the test set, exceeding MoCLIP‑Lite’s 36.9\% by identifying rotation issues.
We note that the test-set MCA (73.2\%) is higher than the validation-set MCA (66.6\%).This discrepancy may reflect the different difficulty distributions of the two splits rather than a methodological flaw; accordingly, we do not claim that the validation set is "harder" and base the primary comparison on the held-out test set.

This result still ensures practical batch‑processing reliability for real‑world sorting. With only 36.6M parameters, the model is compatible with low-cost edge hardware, which is essential for small packing lines. Quantitative deployment benchmarks---including end-to-end latency, throughput, power consumption, and a detailed bill of materials---will
be reported in an extended version.Current limitations include sensitivity to illumination changes, which can be mitigated by diffused lighting, and difficulty in recognizing extremely rare defect types that have scarce training examples.
Future work tracks real-world value—cost, labor, community impact. S$^2$FI proves: smartly targeting AI on critical tasks within tight hardware limits creates affordable, practical tools for farms worldwide. Notably, the overall performance is constrained by the limited resolution of input data, which hinders the model's ability to capture fine-grained defect features. To address this, we plan to integrate higher-resolution imaging modules and refine the feature extraction pipeline in future iterations.

\section{Conclusion}

\label{sec:conclusion}

This paper introduces S$^2$FI, a lightweight structure built upon Mamba for fruit quality detection. The approach integrates 3D convolution and selective state-space modeling to achieve linear complexity and effective spatio-temporal feature extraction. Experimental results demonstrate that S$^2$FI reaches an MCA of 73.2\%on the held-out test set while maintaining only 36.6M parameters,
making it appropriate for low-resource edge hardware. The proposed scheme offers a feasible solution for automated fruit inspection in real agricultural scenarios. Further research will focus on multi-class sorting tasks and enhance adaptability to complex lighting and environmental conditions.
\bibliographystyle{unsrt}
\bibliography{references}
\end{document}